\documentclass{article}
\usepackage[T1]{fontenc}
\usepackage[final]{corl_2026}
\usepackage{amsmath,amssymb}
\usepackage{graphicx}
\hypersetup{
  pdftitle={FailureGen: Scalable Robot Failure Data Generation with World Models},
  pdfauthor={Akul Santhosh, Chaitra Sathe, Aastha Jhunjhunwala},
  pdfsubject={Robot failure data generation with world models},
  pdfkeywords={Synthetic data generation, Robot failure generation, World models, Failure detection}
}
\title{FailureGen: Scalable Robot Failure Data Generation with World Models}
\author{%
\begin{tabular}[t]{@{}c@{\hspace{1.8em}}c@{\hspace{1.8em}}c@{}}
\textbf{Akul Santhosh} & \textbf{Chaitra Sathe} & \textbf{Aastha Jhunjhunwala} \\[3pt]
{\normalfont\small\href{mailto:akuls@nvidia.com}{\texttt{akuls@nvidia.com}}} &
{\normalfont\small\href{mailto:csathe@nvidia.com}{\texttt{csathe@nvidia.com}}} &
{\normalfont\small\href{mailto:aasthaj@nvidia.com}{\texttt{aasthaj@nvidia.com}}} \\[6pt]
\multicolumn{3}{c}{\normalfont NVIDIA} \\[6pt]
\multicolumn{3}{c}{\normalfont\small Project website: \url{https://akul-nv.github.io/failuregen/}}
\end{tabular}}
\date{}
\makeatletter
\renewcommand{\@noticestring}{Accepted to the ``Oops, I erred'' Workshop at CoRL 2026.}
\def\ps@workshop{%
  \def\@oddhead{\hfil\small Accepted to the ``Oops, I erred'' Workshop at CoRL 2026\hfil}%
  \let\@evenhead\@oddhead
  \def\@oddfoot{\hfil\thepage\hfil}%
  \let\@evenfoot\@oddfoot}
\makeatother
\begin{document}
\maketitle
\begin{abstract}
Robot failures are difficult to collect at scale, limiting the range of
failure scenarios represented in robot learning datasets. We present
FailureGen, a method for generating synthetic videos with temporal event
annotations by reusing action trajectories from recorded failures.
The method comprises four stages: fine-tuning an action-conditioned
Cosmos world model on successful and failed real executions; annotating
source recordings to identify failure-associated donor segments and
compatible successful hosts; transferring donor motion into host contexts,
with optional spatial offsets, to generate continuous video rollouts;
and annotating and filtering the generated samples according to observed
events and visual quality. We study two implementations with different
tasks and robot configurations, with visual failure detection serving
as one downstream application. In a preliminary comparison, a detector
trained on real and generated data achieves an average precision of
0.51, compared with 0.41 using real data alone, and detects 33 of 36
failure intervals, compared with 28 of 36. By reusing observed failures
in additional contexts, FailureGen expands the data available for
studying robot failures and developing downstream learning methods.
\end{abstract}

\keywords{Synthetic data generation, Robot failure generation, World models, Failure detection}

\section{Introduction}
\label{sec:introduction}

Collecting diverse examples of robot failures at scale is difficult.
Failures can depend on specific combinations of robot motion, object
configuration, and contact conditions. Capturing this variation requires
repeated physical executions, and less frequent failure modes may still
remain underrepresented.

We present FailureGen, a method for expanding failure datasets by reusing
action trajectories from recorded failures. A failure-associated donor
supplies the motion, while a successful host supplies the target scene
and nominal trajectory. An adapted Cosmos world
model~\cite{nvidia2026cosmos3} generates video rollouts conditioned on
the host scene and transferred motion. Different hosts and optional
spatial offsets vary the context and location of the intervention,
producing additional examples without requiring a new physical execution
for each video.

Prior work has explored synthetic failure construction through procedural
perturbations in simulation, as in AHA~\cite{duan2024aha}, and through
action-conditioned world models with multiple verifiers, as in
Dream2Fix~\cite{li2026dream2fix}. FailureGen offers a complementary
approach by using observed failures as the source of motion interventions
and expanding the contexts in which these behaviors are represented.
This approach makes further use of failures that have already been
captured, including events that would be difficult to collect repeatedly.

Transferred motion may produce a different failure or a successful
continuation in a new scene. Generated videos may also contain visual
artifacts. We therefore label and filter rollouts according to their
observed events and visual quality, producing data for downstream
learning.

Section~\ref{sec:experiments} examines two implementations with different
tasks and robot configurations. We evaluate visual failure detection
as one downstream application through a preliminary comparison of
training on real-only and augmented data.

This paper makes three contributions:
\begin{itemize}
    \item FailureGen, a method for expanding robot failure datasets by
    transferring recorded failure trajectories into compatible host
    scenes and generating video rollouts with an action-conditioned
    world model.

    \item Two experimental implementations applying the method to
    different tasks and robot configurations.

    \item A preliminary evaluation of visual failure detection as a
    downstream application, comparing detectors trained on real-only
    and augmented datasets.
\end{itemize}

\section{Related Work}
\label{sec:related_work}

\paragraph{Simulation and procedural failure generation.}
Simulation provides controlled environments for generating robot failures
and studying their consequences. AHA constructs failure examples through
procedural perturbations for detection and reasoning~\cite{duan2024aha}.
FailSafe pairs generated failures with executable recovery actions,
using simulator execution to validate proposed
corrections~\cite{lin2025failsafe}. DenseReward synthesizes failures such
as missed grasps, collisions, and object drops to learn dense rewards
for downstream planning and policy learning~\cite{fang2026densereward}.
Guardian combines simulated action perturbations with supervision
constructed from real recordings for planning and execution
verification~\cite{pacaud2025guardian}. These methods demonstrate how
targeted changes to robot execution can produce useful examples of
unsuccessful behavior.

\paragraph{World-model-based data generation.}
Action-conditioned world models generate visual continuations from
scene observations and robot motion. Ctrl-World and Cosmos provide
models and adaptation tools for this form of
generation~\cite{guo2025ctrlworld,nvidia2025cosmos,nvidia2026cosmos3},
while RoboDream composes robot motion with scene conditioning for
data synthesis~\cite{ye2026robodream}. DreamGen, VLAW, and WorldSample
use generated experience to support policy learning and
improvement~\cite{jang2025dreamgen,guo2026vlaw,xue2026worldsample}.
For failure generation, Dream2Fix perturbs actions, generates
candidate outcomes with a world model, and applies multiple verifiers
to construct recovery supervision~\cite{li2026dream2fix}.
StressDream steers generated futures toward specified outcomes,
including failures, by optimizing the initial diffusion
noise~\cite{seo2026stressdream}. FailureGen builds on conditional
video generation by supplying motion derived from recorded failures
and transferring it into compatible host contexts.

The reliability of these generated interactions is also an active
research topic. MiraBench evaluates physical consistency, action
following, and optimism bias~\cite{yang2026mirabench}.
WorldSimProbe examines action realization and interaction dynamics,
including a donor--receiver setup that replays actions from one task
in another scene~\cite{co2026worldsimprobe}. This provides a related
use of motion--scene composition for evaluating world models.
Together, these studies motivate assessing generated motion and
interaction outcomes alongside visual quality.

\paragraph{Trajectory reuse and data augmentation.}
Recorded demonstrations provide another basis for expanding robot
datasets. MimicGen transforms demonstrated motion for new object
configurations~\cite{mandlekar2023mimicgen}, while DemoGen adapts
trajectories and synthesizes corresponding point-cloud observations
through spatial transformations~\cite{xue2025demogen}.
Real2Render2Real reconstructs object motion and appearance from
recordings to render robot demonstrations without object-dynamics
simulation~\cite{yu2025real2render2real}. GenAug and ROSIE expand
visual diversity through generative edits to recorded
observations~\cite{chen2023genaug,yu2023rosie}. These approaches
illustrate how existing experience can support additional training
examples with reduced physical data collection.

These lines of work offer complementary ways to expand robot
experience. Capturing the diversity of failures, particularly the
long tail of events across changing scenes and interaction
conditions, remains challenging. FailureGen contributes to this
effort by reusing observed failure motion in additional compatible
contexts. Such motion provides an additional source of interventions
that could be combined with procedural perturbations, visual
augmentation, and guided world-model generation to broaden failure
coverage. The resulting data may support several learning
objectives: prior work studies failure
detection~\cite{grislain2025ifailsense}, diagnosis and corrective
guidance~\cite{zeng2025vifailback}, and reward
learning~\cite{fang2026densereward}. This paper evaluates visual
failure detection as one downstream application.

\section{Method}
\label{sec:method}

\paragraph{Problem setup.}
We are given a source dataset $\mathcal{D}_{\mathrm{src}}$ of successful and
failed robot executions. Each episode contains synchronized visual observations
$\mathbf{o}_{0:N}$ and measured robot states $\mathbf{s}_{0:N}$, where $t$
indexes sampling instants and $N$ is the number of transitions. The subscript
range includes both endpoints. Our goal is to construct an additional dataset
$\mathcal{D}_{\mathrm{gen}}$ of generated video--motion pairs with temporal
event annotations, expanding the contexts in which rare robot failures are
represented. We call a recording that supplies failure-associated motion a
\emph{donor}, and a successful recording that supplies a target scene and
nominal motion a \emph{host}. The generated data can support multiple
downstream learning tasks; in this paper, we evaluate its utility for visual
failure detection.

\begin{figure*}[t]
    \centering
    \includegraphics[width=0.8\textwidth]{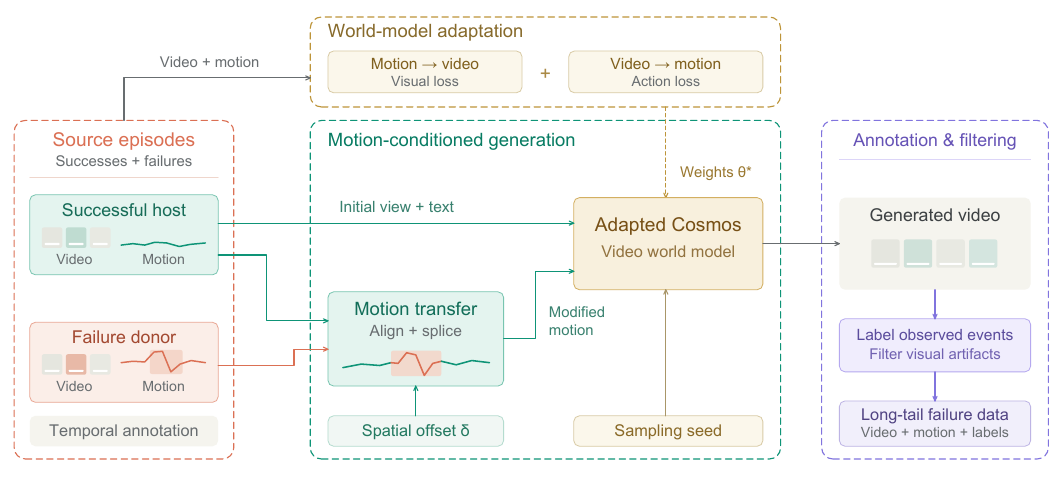}
    \caption{\textbf{FailureGen architecture.}
    Successful and failed source executions support world-model adaptation
    through forward- and inverse-dynamics supervision. At generation time,
    a successful host supplies visual context and nominal motion, while a
    failure donor supplies motion for a selected intervention. Motion transfer
    and optional spatial offsets construct the conditioning sequence for the
    adapted world model. Generated videos are labeled according to their
    observed events, with a cleanup filter for unreliable examples. Each
    retained example includes the video, its associated motion sequence,
    and temporal event labels.}
    \label{fig:failuregen_architecture}
\end{figure*}

As illustrated in Fig.~\ref{fig:failuregen_architecture}, FailureGen comprises
four components: world-model adaptation, source annotation, motion-conditioned
generation, and annotation and filtering of generated data. We adapt the model
using successful and failed executions, identify compatible donor--host
segments, and transfer failure-associated motion into host scenes, optionally
varying the intervention location. Finally, we label the generated events and
filter unreliable examples.

\subsection{World-model adaptation}
\label{sec:adaptation}

We first encode recorded motion into the conditioning representation used by
the world model. For a motion representation $r$, an encoder $\psi_r$ maps
consecutive measured states to a conditioning vector
$\mathbf{a}_t=\psi_r(\mathbf{s}_t,\mathbf{s}_{t+1})$. Joint-space inputs
can use $\Delta\mathbf{q}_t=\mathbf{q}_{t+1}-\mathbf{q}_t$, where
$\mathbf{q}_t$ is the measured joint-position vector. Relative end-effector
inputs can use $\Delta T_t=T_t^{-1}T_{t+1}$, where $T_t$ is a homogeneous
end-effector pose matrix expressed in a fixed reference frame, measured
directly or obtained through forward kinematics. Gripper or other tool states
are included according to the embodiment. The resulting vectors describe
observed motion; measured state transitions need not equal the commands sent
to the robot.

We adapt a pretrained Cosmos world model~\cite{nvidia2026cosmos3} using aligned
video--motion windows from successful and failed executions. We combine
forward- and inverse-dynamics supervision with the aim of improving spatial
dynamics: the visual objective supervises scene evolution under robot motion,
while the action objective supervises motion associated with observed visual
changes.

In forward-dynamics examples, the model generates a visual continuation
conditioned on initial observations, the recorded motion sequence, and a text
description. In inverse-dynamics examples, it predicts motion conditioned on
the observed video. Each training sample is randomly assigned to one of these
two directions, using the model's rectified-flow objective in video-latent or
action space, respectively. Text descriptions provide task or execution
context; each experimental study specifies whether task, phase, or
event-specific captions are used.

Motion enters the model through domain-specific input and output projections,
with a dedicated domain for each embodiment and motion representation.
Fine-tuning updates the generation expert and its associated video, action,
and conditioning modules, while keeping the remaining parameters frozen.
Implementation details and study-specific adaptation settings are provided in
Appendix~\ref{app:method_details}. Motion encoding, coordinate conventions,
visual preprocessing, and normalization remain consistent between adaptation
and generation. We denote the adapted parameters by $\theta^*$ and the forward
model's conditional distribution over visual continuations by $p_{\theta^*}$.

\subsection{Temporal annotation of source data}
\label{sec:real_annotation}

We annotate the source recordings to locate motion segments that can be
transferred. An agentic labeling pipeline using vision-capable LLMs
(Appendix~\ref{app:review_details}) identifies task steps, failures, and
recoveries, assigning each event a category and start and end times. We
distinguish the failure event itself from its visible aftermath, and label
these intervals independently of whether the episode eventually succeeds.
Proposed labels are checked for consistent categories and valid timestamps,
and can be corrected during review.

These annotations guide donor and host selection. We match failure-associated
donor segments with successful host segments at corresponding task steps,
checking compatibility of their robot and object configurations. Recordings
are partitioned by episode before selection, with only training recordings
serving as donors or hosts. Evaluation recordings are also excluded from
world-model adaptation and normalization.

\subsection{Motion-conditioned generation}
\label{sec:generation}

For each selected donor--host pair, we align a failure-associated donor segment
with an intervention window in the host. We index the host window from zero,
let $\tau$ denote the intervention's starting transition within it, and let
$u$ denote the donor segment's starting transition. In addition to transferring
donor motion, we can apply spatial offsets to vary where the intervention
occurs. For a generation horizon of $H$ motion transitions, the transfer
operation constructs the conditioning sequence
\begin{equation}
\widetilde{\mathbf{a}}_{0:H-1}
=
\mathcal{T}_r\!\left(
\mathbf{a}^{h},\mathbf{a}^{d},\mathbf{s}^{h}_{\tau};
\tau,u,\boldsymbol{\delta},\eta
\right).
\end{equation}
Here, superscripts $h$ and $d$ identify host and donor quantities,
$\mathbf{a}^{h}$ and $\mathbf{a}^{d}$ are their encoded motion sequences,
and $\mathbf{s}^{h}_{\tau}$ is the measured host state at intervention onset.
The representation-specific operator $\mathcal{T}_r$ produces the modified
sequence $\widetilde{\mathbf{a}}_{0:H-1}$. The parameters
$\boldsymbol{\delta}$ specify optional spatial offsets or motion-excursion
variations, while $\eta$ specifies the transferred channels, intervention
duration, and transition settings.

The operation anchors donor motion to the host configuration at intervention
onset and applies the requested variation through representation $r$. Each
embodiment specifies the offset coordinate frame and its conversion into
conditioning inputs. Transition settings control how the modified segment
joins the nominal host motion. Unselected channels retain host conditioning
throughout; selected channels retain it outside the intervention and its
transitions. Donor and host recordings must use compatible motion
representations and kinematic conventions.

We then condition the adapted world model on the host's initial observation
$\mathbf{o}^{h}_0$, the modified motion sequence, and a text description $c$:
\begin{equation}
\widehat{\mathbf{o}}_{1:H}\sim
p_{\theta^*}\!\left(
\cdot\mid
\mathbf{o}^{h}_0,\widetilde{\mathbf{a}}_{0:H-1},c
\right).
\end{equation}
The sequence $\widehat{\mathbf{o}}_{1:H}$ contains the $H$ generated future
observations, which follow the host's initial observation to form the generated
clip. Spatial offsets can change where the transferred motion encounters
objects and surfaces. The world model predicts the resulting visual
continuation, allowing the same failure-associated behavior to unfold at
different locations with different contact and object-motion dynamics. The
observed outcome may therefore change even when the donor segment remains
fixed.

We generate candidates by varying compatible donor--host pairs, motion
parameters, and sampling seeds. Each candidate retains its source identifiers,
conditioning sequence, transfer parameters, checkpoint, and sampling settings.
The transferred motion specifies the intended intervention; the generated
video determines the observed event and its timing.

\subsection{Annotation and filtering}
\label{sec:filtering}

We annotate generated videos using the same labeling pipeline as the source
data. Labels describe the events visible in each video, which may differ from
the intended intervention. Because world models can produce visual artifacts,
implausible interactions, or unclear outcomes, a cleanup filter can exclude
clips that cannot provide reliable training examples. The annotation and
filtering policy used for each study is specified in the experiments.

Each retained example contains the generated video, its associated motion
sequence, temporal event labels, and source identifiers. These video--motion
pairs form $\mathcal{D}_{\mathrm{gen}}$ and provide data for downstream learning
tasks.

\section{Experimental Studies}
\label{sec:experiments}

Our experiments examine two aspects of FailureGen: its ability to expand
the available failure data across host contexts, and the usefulness of
these generated examples for downstream learning. We study an in-house
industrial bimanual manipulation task and a focused three-task ViFailback
follow-up. The industrial setting emphasizes multistage execution and
interactions with surrounding equipment, while ViFailback examines
additional manipulation scenarios and downstream model sizes.

Both settings use an adapted Cosmos3-Nano world model with forward- and
inverse-dynamics supervision. Downstream evaluation compares fine-tuning
on real data alone with fine-tuning on the same real data augmented with
FailureGen examples. The industrial experiment additionally includes a
pretrained reference without task-specific fine-tuning.

\paragraph{Visual verification and filtering.}
Generated videos are assessed using the agentic annotation pipeline and
visually checked and verified by human reviewers. Review considers visual
coherence, preservation of the host scene, and the events depicted in each
generation. Unsuitable examples are removed before the data are considered
for downstream use.

Across the accumulated generation experiments, filtering removes
\textbf{9.8\% of the industrial data} and
\textbf{11.2\% of the ViFailback data}, leaving
\textbf{90.2\% and 88.8\%}, respectively, usable.
These cumulative usability rates are reported separately from the
experiment-specific selection of failure clips for downstream training.
Review procedures and filtering details are provided in the appendix.

\subsection{Industrial bimanual manipulation}
\label{sec:study_yam}

\paragraph{Task and data.}
Our in-house dataset captures a bimanual YAM robot performing an industrial
material-handling task: picking up an SMT reel, transferring it to the other
arm, and placing it onto a stack. Successful execution requires coordinated
grasping, handover, and placement. Failures include pickup misses, pickup
slips, and handover drops, which can produce further interactions with the
table, tote, and surrounding objects.

The source collection contains 80 real episodes: 45 clean executions,
12 pickup misses, 13 pickup slips, and 10 handover drops. World-model
adaptation and downstream detector training use the same 66 training
episodes. The remaining 14 recordings provide downstream validation data.
Generator-specific validation and test allocations are detailed in
Appendix~\ref{app:yam_details}.

World-model conditioning uses measured end-effector motion and gripper
state for both arms. Each arm contributes translation, a six-dimensional
rotation representation, and gripper state, yielding a 20-dimensional
conditioning representation.

\paragraph{Failure-data expansion.}
FailureGen transfers failure-associated motion into successful host
recordings, reusing their scene appearance and initial robot--object
configurations. This allows a failure mechanism to unfold in additional
contexts: for example, the consequences of releasing a reel depend on its
position relative to the table and tote.

For the downstream experiment reported here, the annotation pipeline
identifies \textbf{75 synthetic failure clips from 108 donor-transfer
candidates}. These clips span \textbf{26 distinct host recordings and
23 failure donors}. Of the selected clips, 69 reproduce the intended
failure mechanism and six exhibit another observed failure, which
determines their assigned label.

The resulting clips expand the available examples of pickup misses, pickup
slips, and handover drops. Source--generation comparisons examine how the
host scene is preserved while the transferred motion produces different
object trajectories and interactions with surrounding surfaces.

\begin{figure*}[t]
    \centering
    \includegraphics[width=0.8\textwidth]{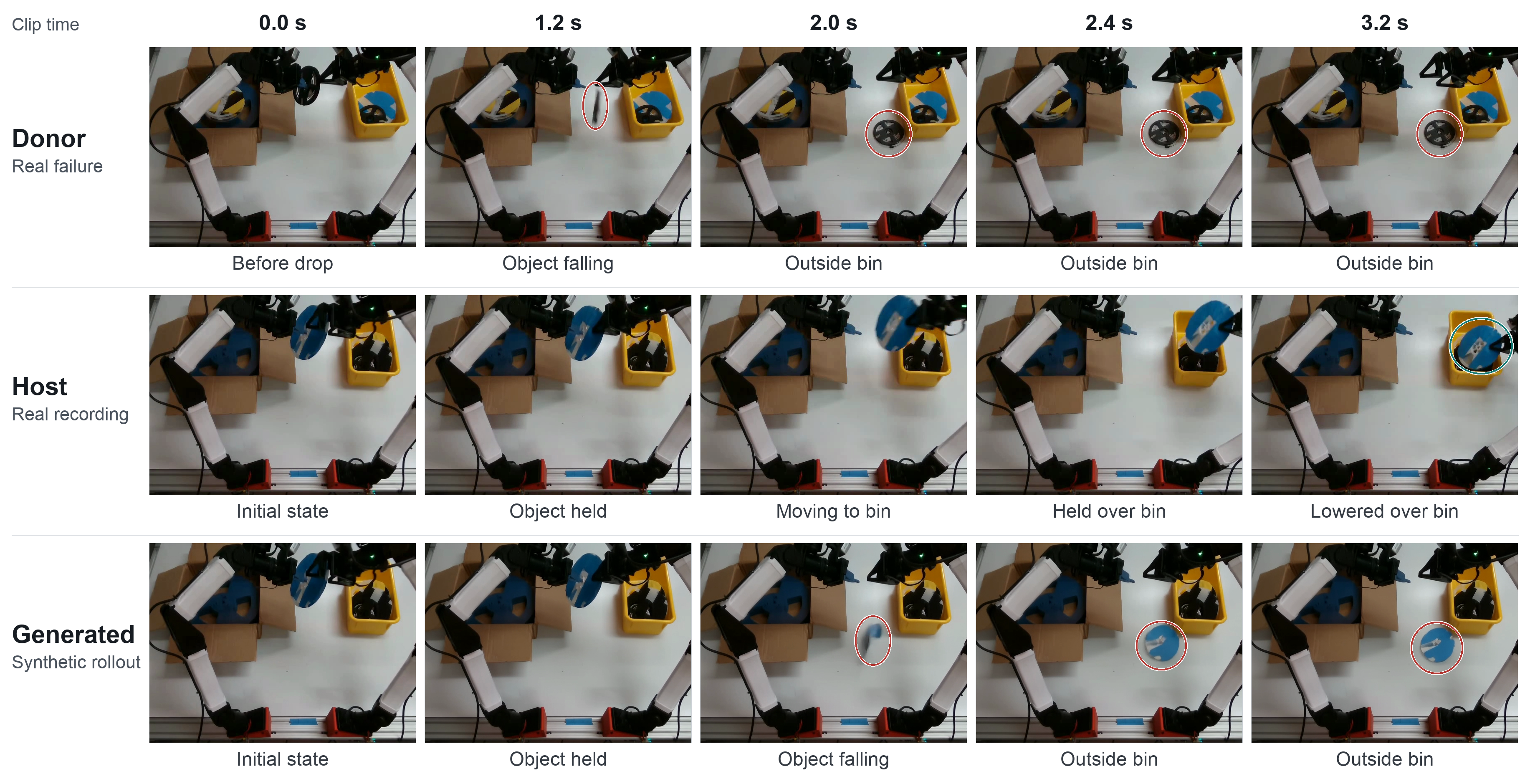}
    \caption{\textbf{Failure transfer in industrial manipulation.}
    Top: a real donor execution in which a reel falls outside
    the bin. Middle: the real host recording, which supplies
    the initial scene and object appearance. Bottom: the
    FailureGen continuation, which preserves the host context
    while reproducing the donor's drop behavior with different
    timing and object motion. Columns show elapsed clip time;
    red outlines highlight the falling or dropped object.}
    \label{fig:industrial_failure_transfer}
\end{figure*}

Figure~\ref{fig:industrial_failure_transfer} illustrates a
donor--host transfer in the industrial task. The donor shows
a reel falling outside the bin, while the host maintains its
grasp during transport. The generated continuation preserves
the host's blue reel and scene arrangement while depicting
a drop onto the surrounding table, transferring the failure
behavior to a different object and initial configuration.

\paragraph{Downstream evaluation.}
We evaluate Cosmos3-Edge-based failure recognition under three conditions:
no task-specific fine-tuning, real-only fine-tuning, and fine-tuning with
FailureGen augmentation. Both fine-tuning conditions use the same 66 real
training episodes; the augmented condition adds the 75 generated failure
clips. Inputs contain video without robot telemetry, and the prediction
target is active failure during execution.

Checkpoints and decision thresholds are selected using real validation
recordings. Evaluation uses a separate real recording session containing
74 eligible episodes: 41 clean episodes and 33 episodes containing
36 annotated failure events. An event is detected if an alert falls within
its annotated interval. We also measure average precision and the number
of clean episodes containing an alert. Model implementation, training
budgets, and evaluation-label review are detailed in
Appendix~\ref{app:yam_details}.

\paragraph{Results.}
The model trained with FailureGen augmentation detects
\textbf{33 of 36 failure events (91.67\%)}, compared with
28 of 36 (77.78\%) for the strongest real-only configuration and
25 of 36 (69.44\%) without fine-tuning
(Table~\ref{tab:industrial_detection}). This corresponds to a
\textbf{13.9-percentage-point increase in event recall} over the strongest
real-only baseline.

The augmented model detects all 12 pickup-slip events and all ten
handover-drop events, together with 11 of 14 pickup misses. It achieves
the highest average precision, 0.51, and raises
\textbf{no alarms in the 41 clean episodes}. The augmented configuration
therefore provides greater failure-event coverage while maintaining
reliable behavior on clean executions.

\begin{table}[t]
    \centering
    \small
    \setlength{\tabcolsep}{4pt}
    \caption{Industrial failure recognition on the shared real evaluation
    subset. Events are detected failure intervals out of 36; clean alarms
    are clean episodes containing an alert out of 41. Real-only reports
    the stronger of the evaluated real-only configurations. The full
    configuration comparison is provided in the appendix.}
    \label{tab:industrial_detection}
    \begin{tabular}{lrrr}
        \hline
        Training condition & Events & Avg Precision & Clean alarms \\
        \hline
        No fine-tuning
            & 25/36 & 0.0412 & 41/41 \\
        Real-only
            & 28/36 & 0.4085 & 5/41 \\
        Real + FailureGen
            & \textbf{33/36} & \textbf{0.5119} & \textbf{0/41} \\
        \hline
    \end{tabular}
\end{table}

\subsection{Focused ViFailback evaluation}
\label{sec:vifailback-study}

\paragraph{Tasks and setup.}
Following an initial task-level screening, we conduct a dedicated follow-up
on three ViFailback tasks~\cite{zeng2025vifailback}: placing an object into
a black box, placing a vial and test tube into rack slots, and opening a
drawer, inserting a can, and closing the drawer. These tasks cover object
placement, precise rack placement, and articulated-object manipulation.
Additional conditioning and evaluation details are provided in
Appendix~\ref{app:vifailback_details}.

The available host recordings vary in object identity, arrangement,
tabletop appearance, clutter, and background configuration. FailureGen
uses joint-motion and gripper-state conditioning to produce additional
examples within these recorded scene contexts. The generation mixture
and source-selection details are provided in the appendix.

For downstream evaluation, Qwen3-VL-2B and Qwen3-VL-8B are fine-tuned on
the focused three-task dataset. For each model, the initial condition uses
real training recordings, while the augmented condition uses the same real
recordings together with task-matched FailureGen examples. Both conditions
are evaluated on the same real evaluation episodes, which are excluded
from training and donor--host selection.

We report episode-level accuracy and failure recall. Accuracy measures
correct success/failure classification across all evaluation episodes;
failure recall measures the fraction of failure episodes correctly
identified.

\begin{figure*}[t]
    \centering
    \includegraphics[width=0.8\textwidth]{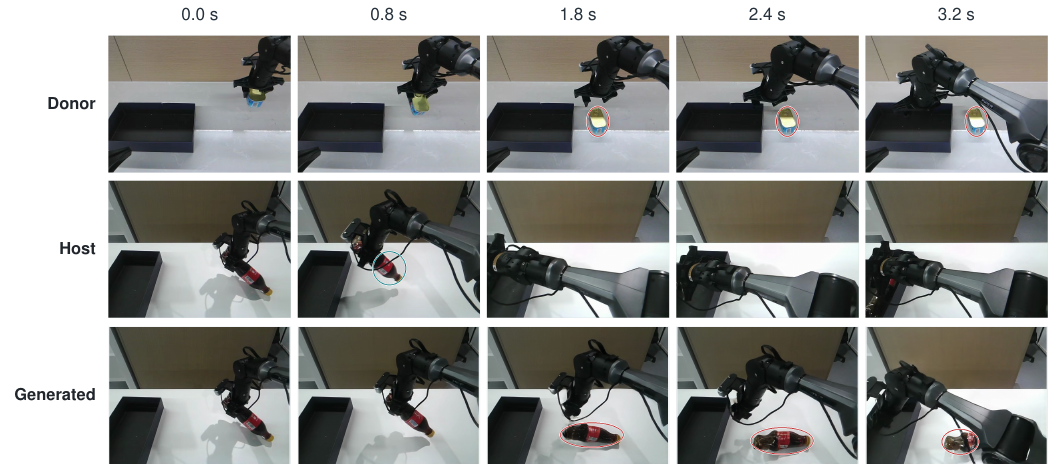}
    \caption{\textbf{Failure transfer for object placement.}
    Top: the donor releases a can outside the box.
    Middle: the host carries a bottle toward the box.
    Bottom: the generated continuation preserves the host's bottle and
    scene while depicting premature release onto the table.
    Times are relative to each clip; red outlines highlight the
    released object.}
    \label{fig:box_failure_transfer}
\end{figure*}

Figure~\ref{fig:box_failure_transfer} illustrates transfer of
premature-release behavior from a can to a bottle while preserving the
host scene. An additional drawer-placement example appears in
Appendix~\ref{app:vifailback_details}
(Fig.~\ref{fig:drawer_failure_transfer}).

\paragraph{Results.}
FailureGen augmentation improves both accuracy and failure recall in
\textbf{five of the six model--task combinations}, with the remaining
combination unchanged
(Table~\ref{tab:vifailback_qwen_initial_augmented}).

The largest improvement occurs in drawer manipulation. Qwen3-VL-2B
accuracy increases from \textbf{84.4\% to 97.0\%}, while failure recall
increases from \textbf{82.5\% to 96.7\%}. Qwen3-VL-8B also improves,
reaching 97.0\% accuracy and 96.7\% failure recall.

For object-into-black-box placement, both models reach
\textbf{100\% accuracy and failure recall} with augmentation.
Qwen3-VL-2B improves from 94.1\% accuracy and 93.3\% recall, while
Qwen3-VL-8B improves from 98.5\% accuracy and 98.3\% recall.

For vial/test-tube rack placement, Qwen3-VL-8B improves from 49.6\%
to 54.1\% accuracy and from 43.3\% to 48.3\% failure recall.
Qwen3-VL-2B remains unchanged. These results extend the observed utility
of FailureGen-augmented training to additional manipulation tasks and
two downstream model sizes.

\begin{table*}[t]
    \centering
    \small
    \setlength{\tabcolsep}{10pt}
    \caption{Failure detection on the three-task ViFailback subset.
    Values show real-only $\rightarrow$ FailureGen-augmented
    performance (\%). Bold values indicate improvements.}
    \label{tab:vifailback_qwen_initial_augmented}
    \begin{tabular}{lcccc}
        \hline
        & \multicolumn{2}{c}{Qwen3-VL-2B}
        & \multicolumn{2}{c}{Qwen3-VL-8B} \\
        \cline{2-3}\cline{4-5}
        Task
        & Accuracy & Failure recall
        & Accuracy & Failure recall \\
        \hline
        Object into black box
        & $94.1 \rightarrow \mathbf{100.0}$
        & $93.3 \rightarrow \mathbf{100.0}$
        & $98.5 \rightarrow \mathbf{100.0}$
        & $98.3 \rightarrow \mathbf{100.0}$ \\
        Vial/test-tube rack
        & $58.5 \rightarrow 58.5$
        & $53.3 \rightarrow 53.3$
        & $49.6 \rightarrow \mathbf{54.1}$
        & $43.3 \rightarrow \mathbf{48.3}$ \\
        Drawer/can
        & $84.4 \rightarrow \mathbf{97.0}$
        & $82.5 \rightarrow \mathbf{96.7}$
        & $94.8 \rightarrow \mathbf{97.0}$
        & $94.2 \rightarrow \mathbf{96.7}$ \\
        \hline
    \end{tabular}
\end{table*}

\section{Conclusion}
\label{sec:conclusion}

We presented FailureGen, a method for expanding long-tail robot
failure data through motion-conditioned world-model generation.
By transferring failure-associated motion into compatible host
contexts, FailureGen produces additional video--motion examples
with varied visual contexts and interactions, annotated according
to their observed outcomes.

Experiments on industrial bimanual manipulation and a focused
ViFailback subset show that these examples can support improved
failure recognition on real recordings. Together, the results
support FailureGen as a way to extend the learning value of
limited failure demonstrations and motivate its evaluation in
broader downstream robotics tasks.
\clearpage
\bibliography{references}
\clearpage
\appendix

\section{Implementation Details}
\label{app:method_details}

\paragraph{World-model adaptation.}
We adapt Cosmos3-Nano through forward- and inverse-dynamics
supervision. Fine-tuning updates the generation expert, time
embedding, video and action projections, and action-modality
embedding; the remaining parameters are frozen. The corresponding
implementation modules are
\texttt{moe\_gen},
\texttt{time\_embedder},
\texttt{vae2llm},
\texttt{llm2vae},
\texttt{action2llm},
\texttt{llm2action}, and
\texttt{action\_modality\_embed}.

\paragraph{Generation inputs.}
Both studies use windows containing 17 observations at 5~Hz
and 16 motion transitions, spanning 3.2~s between the first
and last observations. Motion is derived from measured robot
states at the corresponding sampling instants. Training-derived
normalization is reused during generation. Neutral task
descriptions provide context without specifying a future failure.

\paragraph{Motion variation.}
Donor motion is aligned with the selected host segment, with
transitions connecting the intervention to the host motion.
For ViFailback, variation scales joint-motion excursions around
the host anchor by factors
$\{0.95,0.975,1.025,1.05\}$ while preserving gripper values
and timing. Each generated example retains its donor and host
identifiers, conditioning sequence, transfer settings, and
inference seed.

\section{Industrial Experiment Details}
\label{app:yam_details}

\paragraph{Data and generator.}
The 80 source recordings are partitioned into 66 generator-training,
seven validation, and seven test episodes. The training partition
contains 37 clean executions, ten pickup misses, eleven pickup
slips, and eight handover drops. Downstream detectors use the same
66 training episodes and combine the remaining 14 recordings
for validation. All synthetic donors and hosts come from training.

Generator observations are subsampled from 30 to 5~Hz and assembled
into a $960\times960$ canvas, with the overhead view above the two
wrist views. The 20-dimensional conditioning contains relative
end-effector translation, six-dimensional rotation, and absolute
gripper state for each arm. Poses are obtained through forward
kinematics, with relative motion computed as $T_t^{-1}T_{t+1}$.
Normalization uses training-only first and 99th percentiles.
Equal-probability forward- and inverse-dynamics training uses
batch size 16 for 1,243 updates.

\paragraph{Synthetic supervision.}
The generation cohort pairs each of 27 failure donors with two
hosts and renders each pairing with seeds 0 and 1. The 75 selected
failure clips contain 18 pickup misses, 27 pickup slips, and
30 handover drops. They contribute 128 positive decision targets.
Only active-failure timestamps in synthetic clips receive
supervision; other frames remain available as temporal context.

\paragraph{Detector implementation.}
Fine-tuned detectors initialize independently from Cosmos3-Edge
revision \texttt{2a00e87e} and train its visual tower, projector,
and causal binary prediction heads. Each prediction receives seven
RGB observations at 5~Hz, spanning 1.2~s, on a $640\times320$
three-view canvas. The target indicates active failure at the
final timestamp; unknown labels are excluded. The pretrained
reference retains the language decoder and scores
\emph{yes}/\emph{no} responses under a fixed prompt.

\paragraph{Training and model selection.}
Table~\ref{tab:app_industrial_training} summarizes the fine-tuned
configurations. The stronger real-only baseline in the main paper
uses a 5:1 negative-to-positive target ratio and positive-loss
weight 5.0. Each run is configured for five epochs; validation
average precision selects the epoch-4 checkpoints. Decision
thresholds maximize validation frame F1 and remain fixed during
evaluation. The configurations share their real source episodes
but differ in sampled negatives and optimization budgets.

\begin{table}[t]
    \centering
    \small
    \setlength{\tabcolsep}{4pt}
    \caption{Industrial fine-tuning configurations.
    Update counts describe the configured five-epoch runs.
    Real 1:1 and Real 5:1 denote negative-to-positive sampling
    ratios; Aug.\ denotes real plus FailureGen data.}
    \label{tab:app_industrial_training}
    \begin{tabular}{lrrr}
        \hline
        Quantity & Real 1:1 & Real 5:1 & Aug. \\
        \hline
        Real positive targets & 105 & 105 & 105 \\
        Synthetic positive targets & 0 & 0 & 128 \\
        Real negative targets & 105 & 525 & 233 \\
        Configured updates & 135 & 400 & 295 \\
        \hline
        Events detected & 21/36 & 28/36 & 33/36 \\
        Average precision & 0.3090 & 0.4085 & 0.5119 \\
        Clean episodes with alarms & 4/41 & 5/41 & 0/41 \\
        \hline
    \end{tabular}
\end{table}

\paragraph{Evaluation protocol.}
Testing uses a separate real recording session. The eligible
subset excludes ten clean recordings with an arm outside the
field of view and one transport-drop recording outside the
failure taxonomy. Real evaluation annotations received partial
human review; label corrections and exclusions followed
inspection of false positives. All systems are scored on the
same final subset with fixed weights and validation thresholds.

An event is detected when an alert falls within its labeled
interval. Average precision is computed over decision timestamps,
and a clean-episode alarm occurs when any eligible timestamp
in a clean episode exceeds the threshold. These definitions
distinguish event coverage from frame-level classification.

\section{ViFailback Experiment Details}
\label{app:vifailback_details}

\paragraph{Tasks and conditioning.}
The focused evaluation covers object-into-black-box placement
(task 49), vial/test-tube rack placement (task 55), and
drawer/can manipulation (task 80). Conditioning contains twelve
measured joint-motion channels and two absolute gripper-position
channels. Overhead and wrist observations are subsampled from
25 to 5~Hz. Generation uses the adapted Cosmos3-Nano
exponential-moving-average checkpoint at update 1,500, selected
using validation replay fidelity.

\paragraph{Host variation and generation.}
Hosts provide variation in object identity for black-box placement,
tabletop appearance and setup for rack placement, and background
objects and cabinet pose for drawer manipulation. Generated
examples include donor-motion transfers, host-motion controls,
and original-context failure replays, with labels assigned from
their observed outcomes. Repeated inference seeds provide
additional renders of a source pairing rather than additional
donor--host contexts.

\paragraph{Downstream comparison.}
Qwen3-VL-2B and Qwen3-VL-8B are fine-tuned on the three selected
tasks. Initial models use real training recordings; augmented
models use the same real recordings with task-matched synthetic
examples. Both conditions are evaluated on the same real
recordings, excluded from training and donor--host selection.

Accuracy measures correct episode-level success/failure
classification. Failure recall measures the fraction of failure
episodes identified correctly. These episode-level metrics
are reported separately from the industrial study's
active-event detection metrics.

\begin{figure*}[t]
    \centering
    \includegraphics[width=\textwidth]{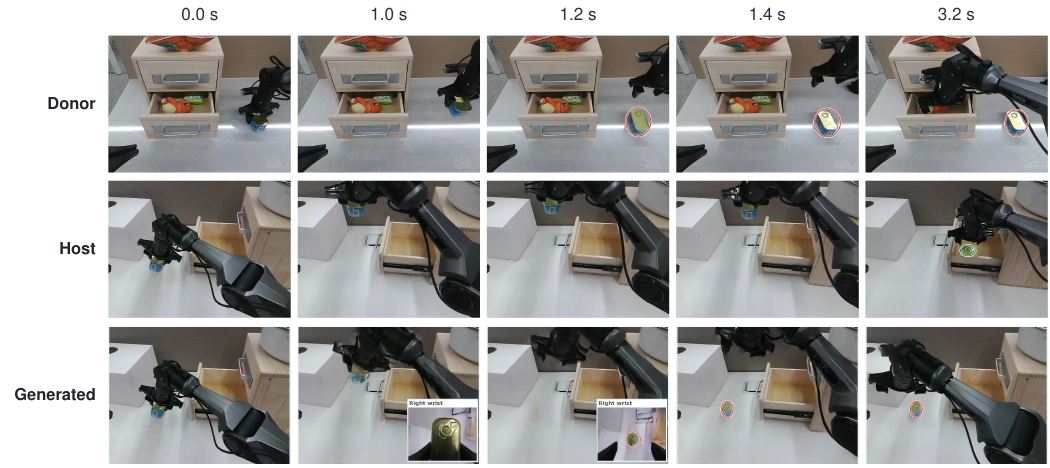}
    \caption{\textbf{Failure transfer for drawer placement.}
    Rows show the real donor, real host, and generated continuation.
    The generated can is released outside the drawer while the host
    scene is retained. Wrist-camera insets show the release obscured
    in the main view. Times are relative to each clip; red outlines
    highlight the released can.}
    \label{fig:drawer_failure_transfer}
\end{figure*}

\paragraph{Additional qualitative example.}
Figure~\ref{fig:drawer_failure_transfer} shows failure transfer in
drawer placement. The host places the can inside the drawer, while the
generated continuation depicts the can falling onto the adjacent tabletop.

Cleanup removes unsuitable generations. Across the accumulated
experiments, removal rates are 9.8\% for industrial manipulation
and 11.2\% for ViFailback, leaving 90.2\% and 88.8\%, respectively,
usable. These usability rates are separate from selecting clips
that contain a particular failure: a coherent successful
continuation can remain usable without becoming a positive
failure-training example.

\section{Agentic temporal annotation}
\label{app:review_details}

The labeling pipeline coordinates vision-capable LLM workers through
evidence preparation, label proposal, and review. Given episode videos and
a task-specific label vocabulary, it prepares chronological frame grids
with synchronized camera views and source-frame indices. Workers inspect
overlapping windows to propose task phases, failure events, and recovery
intervals. Independent review passes check these proposals and look for
missed events.

Candidate events and temporal boundaries receive closer inspection using
denser frame sequences. Disagreements trigger additional visual adjudication,
while unresolved cases are flagged for human review. When available,
robot telemetry can nominate intervals for inspection; event labels are
determined from visual evidence.

Structural validation checks label categories, timestamp bounds, interval
overlaps, and required timeline coverage. The output records event categories,
temporal boundaries, source references, and review status. A review interface
lets human annotators inspect the videos and correct labels or boundaries.
Episode processing runs in parallel, with intermediate results retained
for resuming interrupted annotation.

\end{document}